\def\year{2021}\relax

\documentclass[letterpaper]{article} % DO NOT CHANGE THIS
\usepackage{setspace}
\usepackage{aaai21}  % DO NOT CHANGE THIS
\usepackage{times}  % DO NOT CHANGE THIS
\usepackage{helvet} % DO NOT CHANGE THIS
\usepackage{courier}  % DO NOT CHANGE THIS
\usepackage[hyphens]{url}  % DO NOT CHANGE THIS
\usepackage{graphicx} % DO NOT CHANGE THIS
\usepackage{natbib}  % DO NOT CHANGE THIS AND DO NOT ADD ANY OPTIONS TO IT
\usepackage{caption} % DO NOT CHANGE THIS AND DO NOT ADD ANY OPTIONS TO IT
\usepackage{amsmath}

\usepackage{booktabs}
\usepackage{times}
\usepackage{epsfig}
\usepackage{graphicx,subfigure}
\usepackage{amssymb}
\usepackage{bm}
\usepackage{makecell}
\usepackage{multirow}
\usepackage{wrapfig}
\usepackage{algorithm}
\usepackage{algorithmic}
\usepackage{adjustbox}
\usepackage{bbm}
\usepackage{epstopdf}
\usepackage{threeparttable}
\usepackage{tablefootnote}
\usepackage[normalem]{ulem}

\usepackage{color}
\usepackage[dvipsnames]{xcolor}

\newcommand{\etal}{\textit{et al. }}

\DeclareMathAlphabet{\mathcal}{OMS}{cmsy}{m}{n}

\usepackage{amsmath}
\usepackage{amsthm}

\newcommand{\ie}{\textit{i}.\textit{e}.} 
\usepackage[switch]{lineno}

\newcommand{\name}{DecAug}

\begin{document}

\title{DecAug: Out-of-Distribution Generalization via \\
Decomposed Feature Representation and Semantic Augmentation}
%\author{Paper ID 3667}

\author {
    % Authors
    Haoyue Bai \textsuperscript{\rm 1}\thanks{This work was done while intern at Huawei Noah’s Ark Lab.}\thanks{Equal contribution.},
    Rui Sun \textsuperscript{\rm 2}\footnotemark[2],
    Lanqing Hong \textsuperscript{\rm 2},
    Fengwei Zhou \textsuperscript{\rm 2}, \\
    Nanyang Ye \textsuperscript{\rm 3}\thanks{Corresponding author.},
    Han-Jia Ye \textsuperscript{\rm 4},
    S.-H. Gary Chan \textsuperscript{\rm 1},
    Zhenguo Li \textsuperscript{\rm 2} \\
}
\affiliations {
    % Affiliations
    \textsuperscript{\rm 1} The Hong Kong University of Science and Technology \\
    \textsuperscript{\rm 2} Huawei Noah's Ark Lab \\
    \textsuperscript{\rm 3} Shanghai Jiao Tong University, 
    \textsuperscript{\rm 4} Nanjing University \\
    \{hbaiaa, gchan\}@cse.ust.hk, \{sun.rui3, honglanqing, zhoufengwei, li.zhenguo\}@huawei.com  \\
    ynylincoln@sjtu.edu.cn, yehj@lamda.nju.edu.cn
}

\maketitle

\begin{abstract}
\begin{quote}

While deep learning demonstrates its strong ability to handle independent and identically distributed (IID) data, it often suffers from \emph{out-of-distribution (OoD) generalization},
where the test data come from another distribution (w.r.t. the training one).
Designing a general OoD generalization framework to a wide range of applications is challenging, mainly due to possible correlation shift and diversity shift in the real world. Most of the previous approaches can only solve one specific distribution shift, such as shift across domains or the extrapolation of correlation. To address that, we propose \name{}, a novel {\bf dec}omposed feature representation and semantic {\bf aug}mentation approach for OoD generalization. \name{} disentangles the category-related and context-related features.
Category-related features contain causal information of the target object, while context-related features describe the attributes, styles, backgrounds, or scenes, causing distribution shifts between training and test data. The decomposition is achieved by orthogonalizing the two gradients (w.r.t. intermediate features) of losses for predicting category and context labels. Furthermore, we perform gradient-based augmentation on context-related features to improve the robustness of the learned representations. Experimental results show that \name{} outperforms other state-of-the-art methods on various OoD datasets, which is among the very few methods that can deal with different types of OoD generalization challenges. 
\end{quote}
\end{abstract}

\section{Introduction}
Deep learning has demonstrated superior performances on standard benchmark datasets from various fields, such as image classification \cite{alexnet2012}, object detection \cite{redmon2016you}, natural language processing \cite{bert2019}, and recommendation systems \cite{Cheng2016}, assuming that the training and test data are independent and identically distributed (IID).
In practice, however, it is common to observe distribution shifts among training and test data, which is known as out-of-distribution (OoD) generalization.
How to deal with OoD generalization is still an open problem.

\begin{figure}[!t]
    \centering
    \includegraphics[width=0.65\linewidth]{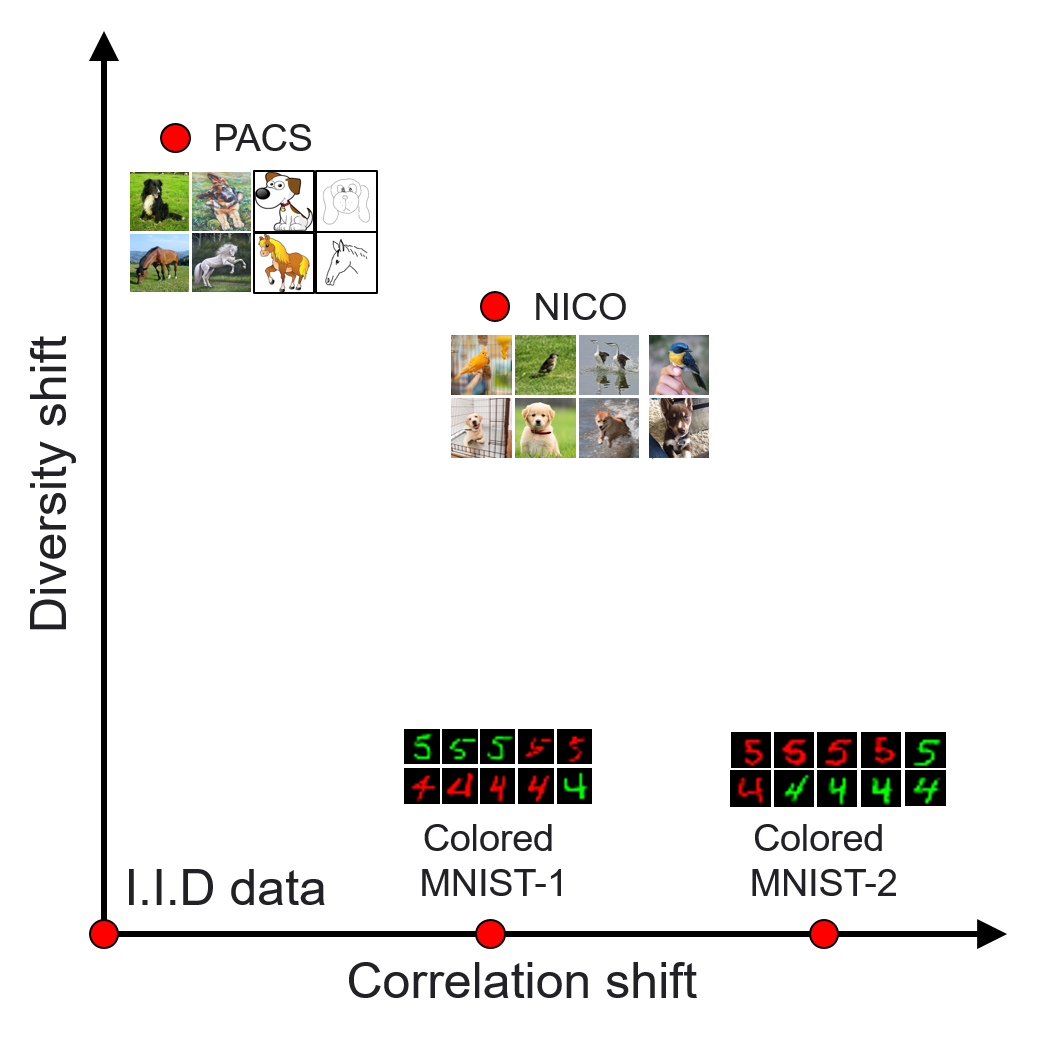}
    \caption{Illustration of the two-dimensional out-of-distribution problem among datasets in different OoD research areas, including Colored MNIST, PACS, and NICO. Extensive experiments showed that many OoD methods can only deal with one dimension of OoD generalization.
    }
    \label{fig:2dimood}
\end{figure}

To improve a DNN's OoD generalization ability, diversified research endeavours are observed recently, which mainly includes domain generalization, invariant risk minimization, and stable learning. Various standard benchmark datasets are adopted to evaluate the proposed OoD generalization algorithms, such as Colored MNIST~\cite{arjovsky2019invariant}, PACS~\cite{Li2017}, and NICO~\cite{he2020towards}. Among these datasets, PACS are widely used in domain generalization~\cite{carlucci2019domain, mancini2020towards} to validate DNN's ability to generalize across different image styles. On the other hand, in recent risk regularization methods, Colored MNIST is often considered \cite{arjovsky2019invariant, ahuja2020invariant, krueger2020outofdistribution}, where digits' colors are either red or green. Manipulating the correlation between the colors and the labels would result in a distribution shift. In stable learning, another OoD dataset called NICO was introduced recently~\cite{he2020towards}.
It contains image style changes, background changes, and correlation extrapolation. Along with this dataset, an OoD learning method, named CNBB, is proposed, based on sample re-weighting inspired by causal inference.

In this paper, we observe that methods perform well in one OoD dataset, such as PACS, which may show very poor performance on another dataset, such as Colored MNIST, as shown in the our experiments. That may because of the different dimensions of OoD generalization.
Here, we identify two types of out-of-distribution factors, including the correlation shift and the diversity shift.

\begin{figure}
    \centering
    \subfigure[Train]{\includegraphics[height = 0.9cm]{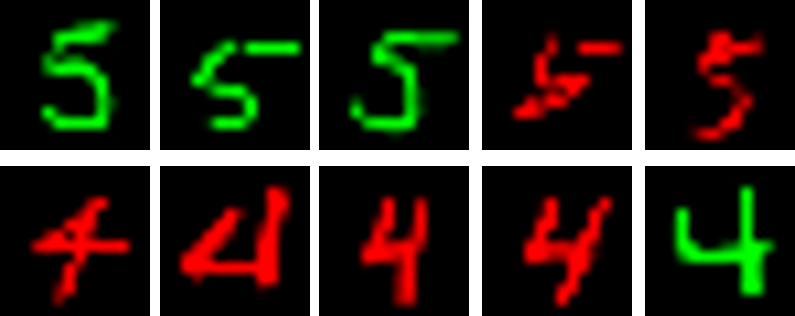}}
    \hspace{1mm}
    \subfigure[Test]{\includegraphics[height = 0.9cm]{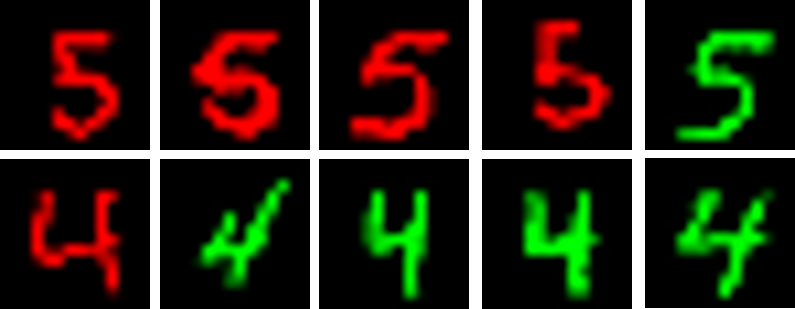}}
    \caption{Typical examples of out-of-distribution correlation shift data from the Colored MNIST dataset.}
    \label{fig:imgcmnist}
\end{figure}

\textbf{Correlation shift.} One is the correlation shift, which means that labels and the environments are correlated and the relations change across different environments. 
For example, in Fig.~\ref{fig:imgcmnist}, we observe the correlation shift between the training set and the test set in Colored MNIST.
Specifically, in the training set, number $5$ is usually in green while number $4$ is usually in red.
However, in the test set, number $5$ tends to be in red while number $4$ tends to be in green.
If a model learns color green to predict label $5$ when training, it would suffer from the correlation shift when testing.

\begin{figure}[t]
    \centering
    \subfigure[P]{\includegraphics[height = 3.0cm]{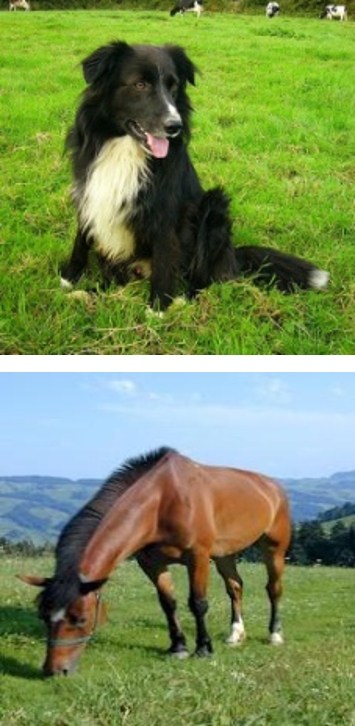}}
    \subfigure[A]{\includegraphics[height = 3.0cm]{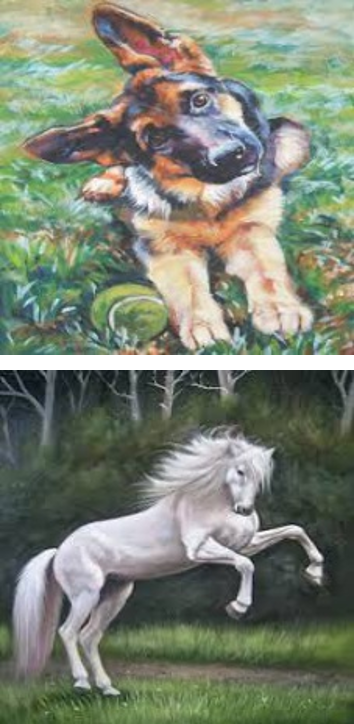}}
    \subfigure[C]{\includegraphics[height = 3.0cm]{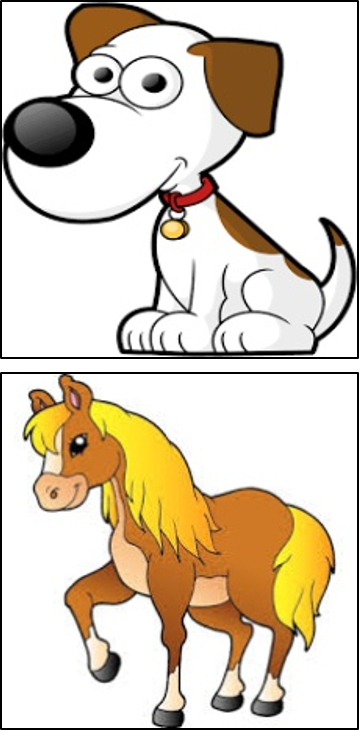}}
    \subfigure[S]{\includegraphics[height = 3.0cm]{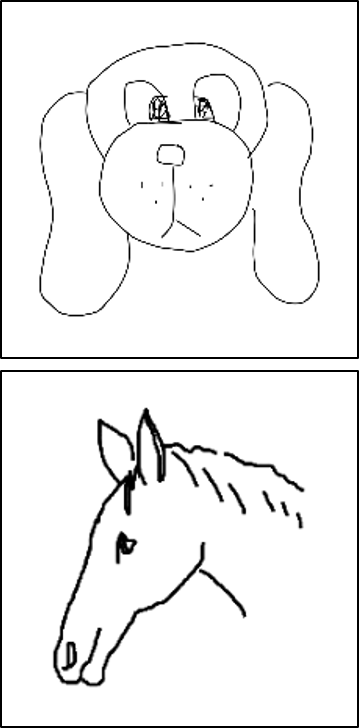}}
    \caption{Typical examples of out-of-distribution diversity shift data from the PACS dataset. (a) Photo. (b) Art Painting. (c) Cartoon. (d) Sketch.}
    \label{fig:imgpacs}
\end{figure}

\textbf{Diversity shift.} Another out-of-distribution factor is the diversity shift. For example, in PACS, the data come from four different domains: photo, art painting, cartoon and sketch. Data in different domains have significantly different styles. Usually, we leave one domain out as the test set, and the remaining three domains as training set. The model trained on the training set would susceptible to the diversity shift on the test set. See Fig.~\ref{fig:imgpacs} as an illustration.

\begin{figure}
    \centering
    \subfigure[In cage]{\includegraphics[height = 3.2cm]{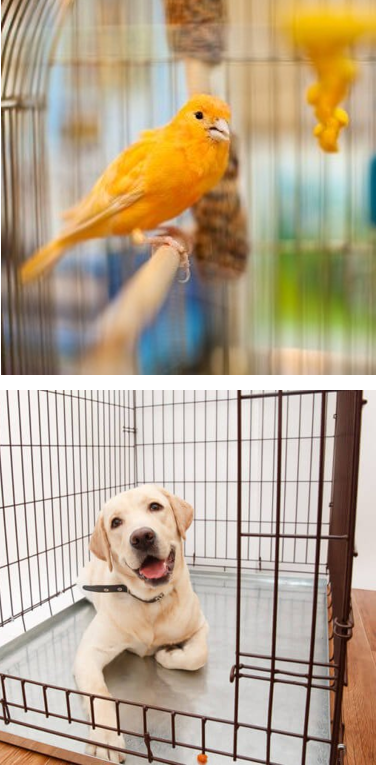}}
    \subfigure[In water]{\includegraphics[height = 3.2cm]{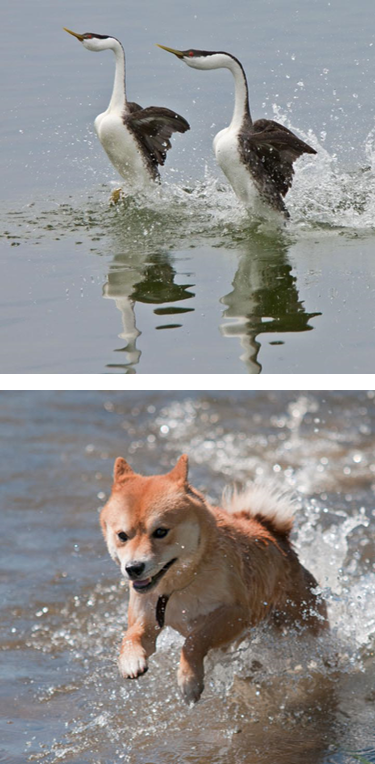}}
    \subfigure[On grass]{\includegraphics[height = 3.2cm]{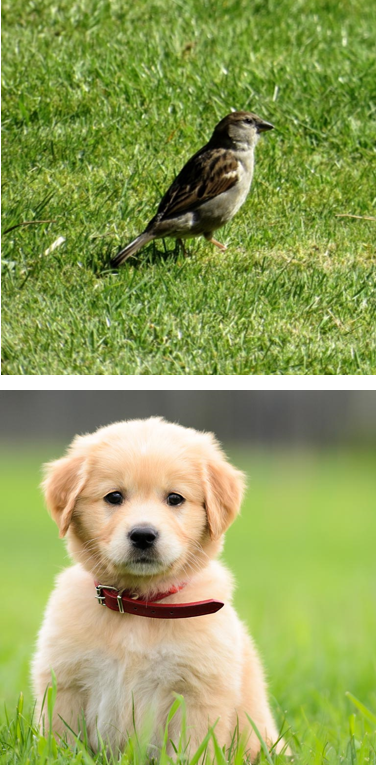}}
    \hspace{1mm}
    \subfigure[Others]{\includegraphics[height = 3.2cm]{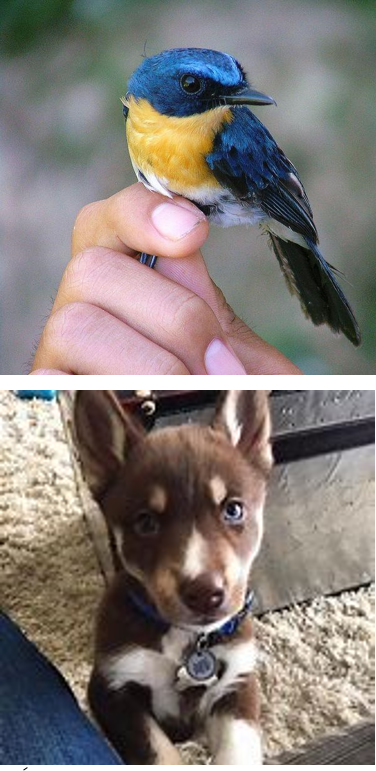}}
    \caption{Some examples of the two-dimensional out-of-distribution data from the NICO dataset.
    }
    \label{fig:imgnico}
\end{figure}

\textbf{Two-dimension OoD.} Data in actual scenarios usually involve two different OoD factors simultaneously.
For example, in NICO (Fig.~\ref{fig:imgnico}), different contexts such as ``in cage", ``in water", and ``on grass" lead to diversity shift, while some contexts are related to specific categories, such as a bird would be ``in hand'' and a dog may be ``at home''.
We also put datasets from multiple research areas on the same axis (Fig.~\ref{fig:2dimood}), the $X$-axis denotes the correlation shift which controls the contribution proportions of correlated features, the $Y$-axis denotes the diversity shift which stands for the change of feature types. Specifically, in the Colored MNIST dataset, the correlation between color and label is high, while in the PACS, the style of images is more diverse. In the NICO, both correlation shift and diversity shift exist.

In our experiments, we showed that algorithms considered only one-dimension of OoD generalization may fail in the other dimension of OoD. To handle different OoD factors simultaneously, we propose DecAug, a novel decomposed feature representation and semantic augmentation approach for OoD generalization. Specifically, our method first decomposes the high-level representations of input images into category-related and context-related features by orthogonalizing the two gradients of losses for predicting category and context labels respectively. Category-related features are essential for recognizing the category labels of the images, while context-related features are not essential for the recognition but correlated with the category labels. After obtaining the decomposed features, we do gradient-based semantic augmentation on context-related features, representing attributes, styles, backgrounds, or scenes of target objects, to disentangle the spurious correlation between features that are not essential for the recognition and category labels. 

Our contributions are as follows:
\begin{enumerate}
    \item We test OoD methods from diversified research areas and show that very often, they only deal with one special type of OoD generalization challenge. 
    \item We propose \name{} to learn disentangled features that capture the information of category and context respectively and perform gradient-based semantic augmentation to enhance the generalization ability of the model. This is the first data augmentation method that attempts to deal with various OoD generalization problems indicating a new promising direction for OoD generalization.
    \item Extensive experiments show that our method consistently outperforms previous OoD methods on various types of OoD tasks. For instance, we achieve an average accuracy of 82.39\% with ResNet-18~\cite{he2016deep} on PACS~\cite{Li2017}, which is the state-of-the-art performance.
\end{enumerate}

\section{Related Work}
\label{sec:rela}
In this section, we review literature related to risk regularization methods, domain generalization, stable learning, data augmentation and disentangled representation.

\noindent\textbf{Risk regularization methods for OoD generalization.} The invariant risk minimization (IRM, \cite{arjovsky2019invariant}) is motivated by the theory of causality and causal Bayesian networks (CBNs), aiming to find an invariant representation of data from different training environments. To make the model robust to unseen interventions, the invariant risk minimization added invariant risk regularization to monitor the optimality of a dummy classifier on different environments. IRM-Games~\cite{ahuja2020invariant} further improves the stability of IRM. Risk extrapolation (Rex, \cite{krueger2020outofdistribution}) adopts a min-max framework to derive a model that can perform well on the worst linear combination of risks from different environments. These methods typically perform well on synthetic datasets, such as Colored MNIST. However, it is unknown how they can generalize on more complex practical datasets beyond MNIST classification tasks.

\noindent\textbf{Domain generalization.} 
Fabio \etal proposed a self-supervised learning method---Jigsaw that achieves good performance in typical domain generalization datasets, such as PACS \cite{carlucci2019domain}.  A subnetwork sharing weights with the main network is used to solve Jigsaw puzzles. This self-supervised learning method helps in improving the generalization of model on unseen domains. Qi \etal adopted meta learning to learn invariant feature representations across domains \cite{dou2019domain}. Recently, Mancini \etal proposed the curriculum mixup method for domain generalization, in which data from multiple domains in the training dataset mix together by a curriculum schedule of mixup method \cite{mancini2020towards}. Domain generalization methods have achieved performance gain in generalizing models to unseen domains. However, recent OoD research finds that domain adaptation methods with similar design principles can have problems when training distribution is largely different from test distribution \cite{arjovsky2019invariant}.

\noindent\textbf{Stable learning.} Stable learning is a recently proposed new concept \cite{Kuang2018}, which focuses on learning a model that can achieve stable performances across different environments. The methodology of stable learning largely inherited from sampler reweighting in causal inference \cite{Kuang2018, shen2019stable, he2020towards}. For example, in the convnets with batch balancing algorithm (CNBB), instead of viewing all samples in the dataset equally, it first calculates the weights of each sample by a confounder balancing loss which tests whether including or excluding a sample's feature can lead to a significant change in the latent feature representations to reduce the effects of samples that are largely affected by confounders \cite{Kosuke2014,Robins1994}. While these methods can have theoretical guarantees on simplified models, when confounder results in strong spurious correlations, this method may not be able to work well especially in the deep learning paradigm.

\begin{figure*}
    \centering
    \includegraphics[width=0.82\linewidth]{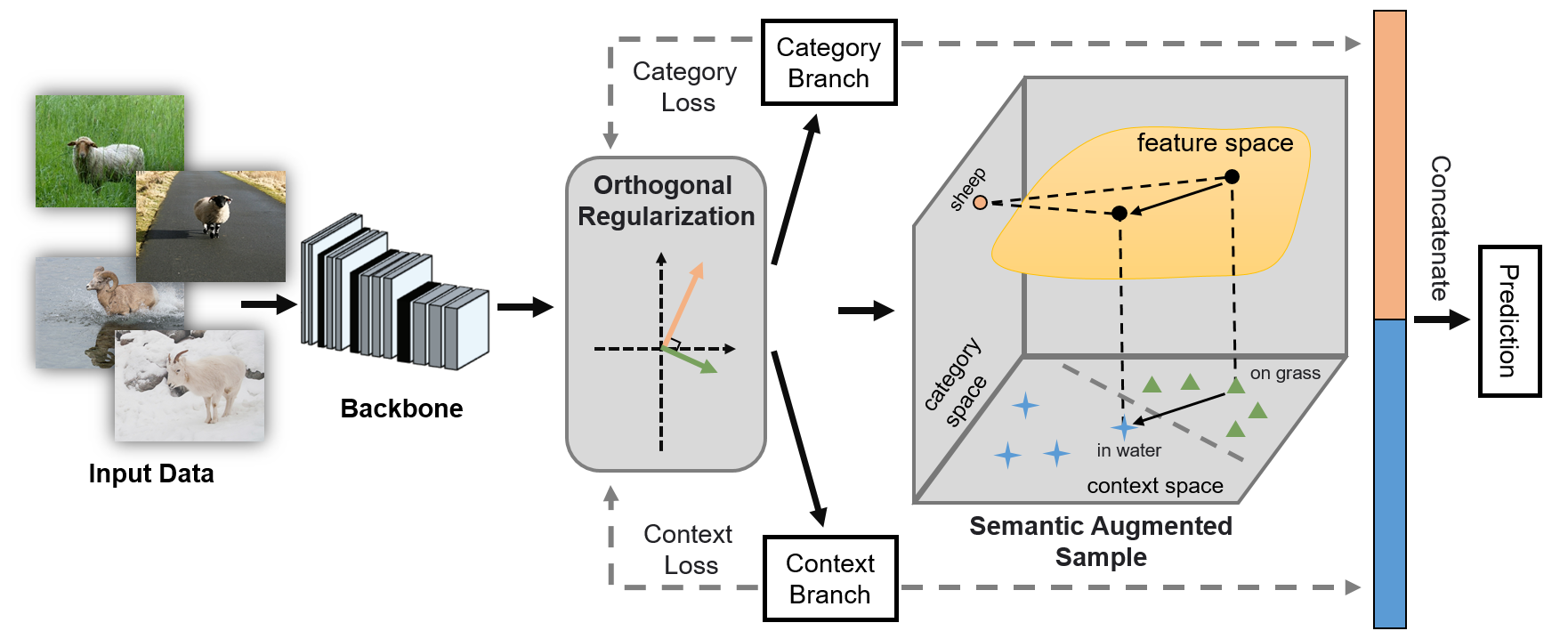}
    \caption{An overview of the proposed \name{}. The input features $z$ extracted by the backbone are decomposed into category-related and context-related features with orthogonal regularization. Gradient-based augmentation is then preformed in the feature space to get semantic augmented samples.}
    \label{fig:framework}
\end{figure*}

\noindent\textbf{Data augmentation.} Data augmentation has been widely used in deep learning to improve the generalization ability of deep models~\cite{alexnet2012,srivastava2015training,han2017deep}. Elaborately designed augmentation strategies, such as Cutout~\cite{devries2017improved}, Mixup~\cite{zhang2017mixup}, CutMix~\cite{yun2019cutmix}, and AugMix~\cite{hendrycks2019augmix}, have effectively improved the performance of deep models. A more related augmentation method is to interpolate high-level
representations. \citet{upchurch2017deep} shows that simple linear interpolation can achieve meaningful semantic transformations. Motivated by this observation, \citet{wang2019implicit} proposes to augment deep features with random vectors sampled from class-specific normal distributions. Instead of augmenting the features explicitly, they minimize an upper bound of the expected loss on augmented data. To tackle few-shot learning problem, \citet{hariharan2017low} suggest to train a feature generator that can transfer modes of variation from categories of a large dataset to novel classes with limited samples. To ease the learning from long-tailed data, \citet{liu2020deep} proposes to transfer the intra-class distribution of head classes to tail classes by augmenting deep features of instances in tail classes. Different from these approaches, our method performs gradient-based augmentation on disentangled context-related features to eliminate distribution shifts for various OoD tasks.

\noindent\textbf{Disentangled representation.} Disentangling the latent factors from the image variants is a promising way to provide an understanding of the observed data \cite{chen2016infogan,higgins2017beta,ma2019disentangled}.
It aims to learn representations that separate the explanatory factors of variations behind the data. Such representations are more resilient to the complex variants and able to bring enhanced generalization ability \cite{liu2018detach,peng2019domain}.
Disentangled representations are inherently more interpretable.
How to obtain disentanglement is still a challenging problem.
\citet{shen2020closed} identifies latent semantics and examines the representation learned by GANs. It derives a closed-form factorization method to discover latent semantic and prove that all semantic directions found are orthogonal in the latent space.
\citet{bahng2019rebias} trains a de-biased
representation by encouraging it to be different from a set of representations that are biased by design.
The method discourages models from taking bias shortcuts, resulting in improved performances on de-biased test data.
In this paper, semantic vectors found by DecAug with orthogonal constraints are disentangled from each other in the feature space.

\section{Methodology}
\label{sec:meth}

To deal with the aforementioned two different types of distribution shifts simultaneously, we argue that it is critical to obtain the decomposed features, one is essential for predicting category and the other is not essential but correlated for recognition. In this section, we propose \name{} to learn decomposed high-level representations for the input data. The decomposition is achieved by orthogonalizing the two gradients of losses for predicting category and context labels respectively. To improve the generalization ability, we perform gradient-based semantic augmentation on context-related features and concatenate the augmented features to category-related features to make the final prediction. An overview of the proposed method is illustrated in Fig.~\ref{fig:framework}.

\subsection{Feature Decomposition} 

Consider an image recognition task with the training set $\mathcal{D} = \{ (x_i, y_i, c_i) \}_{i=1}^{N}$, where $x_i$ is the input image, $y_i$ is the corresponding category label, $c_i$ is the corresponding context label, and $N$ is the number of training data. As shown in Fig.~\ref{fig:framework}, the input data are mapped to the feature space and are decomposed into two branches: category branch and context branch. Given an input image $x_i$ with category label $y_i$ and context label $c_i$, let $z_i=g_{\theta}(x_i)$ be the features extracted by a backbone $g_{\theta}$. For the category branch, $z_i$ is decomposed into $z^1_i = f_{\theta^1}(z_i)$ by a category feature extractor $f_{\theta^1}$, followed by a classifier $h_{\phi^1}(z^1_i)$ to predict the category label. For the context branch, $z_i$ is decomposed into $z^2_i = f_{\theta^2}(z_i)$ by a context feature extractor $f_{\theta^2}$, followed by a classifier $h_{\phi^2}(z^2_i)$ to predict the context label. We use the standard cross-entropy losses $\mathcal{L}^1_i(\theta, \theta^1, \phi^1) = \ell(h_{\phi^1}\circ f_{\theta^1}(z_i), y_i)$ and $\mathcal{L}^2_i(\theta, \theta^2, \phi^2) = \ell(h_{\phi^2}\circ f_{\theta^2}(z_i), c_i)$ to optimize these two branches, together with the backbone, respectively.

\begin{algorithm}[!ht]\small
\caption{\name{}: Decomposed Feature Representation and Semantic Augmentation for OoD generalization}
\label{alg:method}
\begin{algorithmic}[1]
\REQUIRE Training set $\mathcal{D}$, batch size $n$, learning rate $\beta$, hyper-parameters $\epsilon$, $\lambda^1$, $\lambda^2$, $\lambda^{\text{orth}}$.
\ENSURE $\theta$, $\theta^1$, $\phi^1$, $\theta^2$, $\phi^2$, $\phi$.
\STATE Initialize $\theta$, $\theta^1$, $\phi^1$, $\theta^2$, $\phi^2$, $\phi$;
\REPEAT
\STATE Sample a mini-batch of training images $\{(x_i, y_i, c_i)\}_{i=1}^{n}$ with batch size $n$;
\FOR{each $(x_i, y_i, c_i)$}
\STATE $z_i \leftarrow g_{\theta}(x_i)$;
\STATE $\mathcal{L}^1_i(\theta, \theta^1, \phi^1) \leftarrow \ell(h_{\phi^1}\circ f_{\theta^1}(z_i), y_i)$;
\STATE $\mathcal{L}^2_i(\theta, \theta^2, \phi^2) \leftarrow \ell(h_{\phi^2}\circ f_{\theta^2}(z_i), c_i)$;
\STATE Compute $\mathcal{L}^{\text{orth}}_i(\theta^1, \phi^1, \theta^2, \phi^2)$ according to Eq.~\eqref{equ:orthloss};
\STATE Randomly sample $\alpha_i$ from $[0,1]$;
\STATE Generate $\tilde{z}^2_i$ according to Eq.~\eqref{equ:featureaug};
\STATE $\mathcal{L}^{\text{concat}}_i(\theta, \theta^1, \theta^2, \phi) \leftarrow \ell(h_{\phi}([f_{\theta^1}(z_i), \tilde{z}^2_i]),y_i)$;
\STATE Compute $\mathcal{L}_i(\theta, \theta^1, \phi^1, \theta^2, \phi^2, \phi)$ according to Eq.~\eqref{equ:finalloss};
\ENDFOR
\STATE $(\theta, \theta^1, \phi^1, \theta^2, \phi^2, \phi) \leftarrow (\theta, \theta^1, \phi^1, \theta^2, \phi^2, \phi)$ \\ \rightline{$- \beta \cdot \nabla \frac{1}{n} \sum\limits_{i=1}^n \mathcal{L}_i(\theta, \theta^1, \phi^1, \theta^2, \phi^2, \phi)$;}
\UNTIL convergence;
\end{algorithmic}
\end{algorithm}

It is known that the direction of the non-zero gradient of a function is the direction in which the function increases most quickly, while the direction that is orthogonal to the gradient direction is the direction in which the function increases most slowly. To better decompose the features into category-related and context-related features, we enforce the gradient of the category loss $\ell(h_{\phi^1}\circ f_{\theta^1}(z_i), y_i)$ to be orthogonal to the gradient of the context loss $\ell(h_{\phi^2}\circ f_{\theta^2}(z_i), c_i)$ with respect to $z_i$, such that the direction that changes the category loss most quickly will not change the context loss from $z_i$ and vice versa. Specifically, let $\mathcal{G}^1_i(\theta^1, \phi^1) = \nabla_{z_i} \ell(h_{\phi^1}\circ f_{\theta^1}(z_i), y_i)$ and $\mathcal{G}^2_i(\theta^2, \phi^2) = \nabla_{z_i} \ell(h_{\phi^2}\circ f_{\theta^2}(z_i), c_i)$ be the gradients of the category and context loss with respect to $z_i$ respectively. To ensure the orthogonality, we minimize the following loss:
\small
\begin{equation}\label{equ:orthloss}
\mathcal{L}^{\text{orth}}_i(\theta^1, \phi^1, \theta^2, \phi^2) = (\frac{\mathcal{G}^1_i(\theta^1, \phi^1)}{\left\lVert \mathcal{G}^1_i(\theta^1, \phi^1) \right\rVert} \cdot \frac{\mathcal{G}^2_i(\theta^2, \phi^2)}{\left\lVert \mathcal{G}^2_i(\theta^2, \phi^2) \right\rVert})^2.
\end{equation}
\normalsize

\subsection{Semantic Augmentation}
Considering that context-related features cause correlation or diversity shifts in our setting, we perform augmentation on the context-related features to eliminate such kind of distribution shifts. As in the semantic feature space, we may have multiple alternative directions for OoD. To ensure good performances across different environments, we postulate a worse case for the model to learn for OoD generalization by calculating the adversarially perturbed examples in the feature space.
Specifically, let $\mathcal{G}^{\text{aug}}_i = \nabla_{z^2_i} \ell(h_{\phi^2}(z^2_i), c_i)$ be the gradient of the context loss with respect to $z^2_i$. We augment the context-related features $z^2_i$ as follows:
\small
\begin{equation}\label{equ:featureaug}
\tilde{z}^2_i = z^2_i + \alpha_i \cdot \epsilon \cdot \frac{\mathcal{G}^{\text{aug}}_i}{\left\lVert \mathcal{G}^{\text{aug}}_i \right\rVert},
\end{equation}
\normalsize
where $\epsilon$ is a hyper-parameter that determines the maximum length of the augmentation vectors and $\alpha_i$ is randomly sampled from $[0,1]$.

After augmenting the context-related features, we concatenate $\tilde{z}^2_i$ to the category-related features $z^1_i$ to make the final prediction $h_{\phi}([z^1_i, \tilde{z}^2_i])$, where $h_{\phi}$ is a classifier and $[z^1_i, \tilde{z}^2_i]$ is the concatenation of two features. We still use the standard cross-entropy loss $\mathcal{L}^{\text{concat}}_i(\theta, \theta^1, \theta^2, \phi) = \ell(h_{\phi}([z^1_i, \tilde{z}^2_i]),y_i)$ to optimize the corresponding parameters. Together with the aforementioned losses, the final loss is then defined as
\small
\begin{equation}\label{equ:finalloss}
\begin{aligned}
\mathcal{L}_i(\theta, \theta^1, \phi^1, \theta^2, \phi^2, \phi) &= \mathcal{L}^{\text{concat}}_i(\theta, \theta^1, \theta^2, \phi) \\
&+ \lambda^1 \cdot \mathcal{L}^1_i(\theta, \theta^1, \phi^1) + \lambda^2 \cdot \mathcal{L}^2_i(\theta, \theta^2, \phi^2)\\
&+ \lambda^{\text{orth}} \cdot \mathcal{L}^{\text{orth}}_i(\theta^1, \phi^1, \theta^2, \phi^2),
\end{aligned}
\end{equation}
\normalsize
where $\lambda^1$, $\lambda^2$ and $\lambda^{\text{orth}}$ are hyper-parameters that balance different losses. We formulate the learning of DecAug as the following optimization problem:
\small
\begin{equation}\label{equ:objective}
\min_{\theta, \theta^1, \phi^1, \theta^2, \phi^2, \phi}\, \frac{1}{N}\sum_{i=1}^{N} \mathcal{L}_i(\theta, \theta^1, \phi^1, \theta^2, \phi^2, \phi).
\end{equation}
\normalsize
The stochastic gradient descent (SGD) algorithm can be applied to optimize the above objective. The detailed procedures are summarized in Algorithm~\ref{alg:method}.

\section{Experiments}
\label{sec:expe}

In this section, we will conduct numerical experiments to cross benchmark different methods from different perspective of OoD research on different typically challenging and widely used dataset--Colored MNIST, NICO, and PACS.

\subsection{Implementation Details and Datasets}

We evaluate our method on three challenging OoD datasets with different levels of correlation shift and diversity shift as discussed above: Colored MNIST~\cite{arjovsky2019invariant}, PACS~\cite{Li2017}, and NICO~\cite{he2020towards}. The main task of DecAug is category classification subject to unseen data distributions. For PACS and Colored MNIST, the context labels are domain/environment IDs. For NICO, it is attributes. The metric is the top-1 category classification accuracy.

\noindent \textbf{The Colored MNIST Dataset.} The challenging Colored MNIST dataset was recently proposed by IRM~\cite{arjovsky2019invariant} via modifying the original MNIST dataset with three steps: 1) The original digits ranging from 0 to 4 were relabelled as 0 and the digits ranging from 5 to 9 were tagged as 1; 2) The labels of 0 have a  probability of $25\%$ to flip to 1, and vice versa; 3) The digits were colored either red or green based on different correlation with the labels to construct different environments (e.g., $80\%$ and $90\%$ for the training environments and $10\%$ for the test environment). In this way, the classifiers will easily over-fit to the spurious feature (e.g., color) in the training environments and ignore the shape feature of the digits.

For a fair comparison, we followed the same experimental protocol as in IRM~\cite{arjovsky2019invariant} on the Colored MNIST dataset. We equipped the IRMv1 scheme with our \name{} approach using the same settings. The backbone network was a three-layer MLP. The total training epoch was 500 and the batch size was the whole training data. We used the SGD optimizer with an initial learning rate of 0.1. The trained model was tested at the final epoch.

\noindent \textbf{The PACS Dataset.} This dataset contains 4 domains (Photo, Art Painting, Cartoon, Sketch) with 7 common categories (dog,
elephant, giraffe, guitar, horse, house, person). We followed the same leave-one-domain-out validation experimental protocol as in ~\cite{Li2017}. For each time, we select three environments for training and the remaining environment for testing.

The backbone network we used on PACS dataset was ResNet-18. We followed the same training, validation and test split as in JiGen~\cite{carlucci2019domain}. The number of training epochs was 100. The batch size was 64. We used the SGD optimizer with a learning rate of 0.02.

\noindent \textbf{The NICO Dataset.} This dataset contains 19 classes with 9 or 10 different contexts, i.e., different object poses, positions, backgrounds, and movement patterns, etc. The NICO dataset is one of the newly proposed OoD generalization benchmark in the real scenarios~\cite{he2020towards}. The contexts in validation and test set will not appear in training set. 

The backbone network was ResNet-18 without pretraining on the NICO dataset. The number of training epochs was 500 and the batch size was 128. We used the SGD optimizer with a learning rate of 0.05.

We compare our proposed \name{} with the state-of-the-arts, including empirical risk minimization (ERM), invariant risk minimization (IRM, \cite{arjovsky2019invariant}), invariant risk minimization games (IRM-Games, \cite{ahuja2020invariant}), model-agnostic learning of semantic features (MASF, \cite{dou2019domain}), domain generalization by solving jigsaw puzzles (JiGen, \cite{carlucci2019domain}) across multiple datasets, debiased training method (ReBias, \cite{bahng2019rebias}), risk extrapolation (Rex, \cite{krueger2020outofdistribution}), and convnets with batch balancing (CNBB, \cite{he2020towards}).

Our framework was implemented with PyTorch 1.1.0, CUDA v9.0.
For the baseline methods, we implement either with Pytorch 1.1.0 or with Tensorflow 1.8 to keep the same setting as their original source code. IRM, JiGen, ReBias, Rex, and CNBB\footnote{The code of CNBB is from the authors of the paper.} were implemented with Pytorch. IRM-Games and MASF were implemented with Tensorflow.
We conducted experiments on NVIDIA Tesla V100. More implementation details can be found in the Appendix.

\subsection{Results and Discussion}
\label{exp:results}
In this section, we evaluate and analyze the results of our approach on three datasets: Colored MNIST, PACS and NICO. These datasets represent different aspects of covariant shifts in OoD problems thus provide more thorough studies on OoD generalization compared with previous ones.

\begin{table*}[!t]
    \centering
    \caption{Classification accuracy of our approach trained considering leave-one-domain-out validation compared with the state-of-the-art methods on the PACS benchmark with the ResNet-18 backbone.}
    \label{table:pacs}
	\vspace{-0.3cm}
	\begin{adjustbox}{max width=0.8\textwidth}
	\begin{threeparttable}
        \begin{tabular}{lcccc|c}
        \toprule
        \toprule
         Model  &Art Painting  &Cartoon  &Sketch  &Photo  &Average\\
        \midrule
        ERM~\cite{carlucci2019domain}    &77.85 	   &74.86    &67.74   &95.73  &79.05\\
        IRM~\cite{arjovsky2019invariant}\tnote{*}    &70.31 	       &73.12    &75.51   &84.73  &75.92\\
        Rex~\cite{krueger2020outofdistribution}\tnote{*} &76.22&73.76&66.00&95.21&77.80\\
        JiGen~\cite{carlucci2019domain}   &  79.42 & 75.25 &71.35& \textbf{96.03} &  80.51\\
        DANN~\cite{ganin2016domain}      &81.30         &73.80    &74.30   &94.00  &80.80\\
        MLDG~\cite{li2017learning}      &79.50         &77.30    &71.50   &94.30  &80.70\\
        CrossGrad~\cite{shankar2018generalizing} &78.70         &73.30    &65.10   &94.00  &80.70\\
        MASF~\cite{dou2019domain}  &  80.29  & 77.17	& 71.69 & 94.99  & 81.03\\
        Cumix~\cite{mancini2020towards}  &  \textbf{82.30}	 & 76.50 & 	72.60 & 95.10 &  81.60\\
        \midrule
        \emph{DecAug}   & \emph{79.00}	 & \textbf{\emph{79.61}} & 	\textbf{\emph{75.64}} & \emph{95.33} &  \textbf{\emph{82.39}}\\
        \bottomrule
        \bottomrule
        \end{tabular}
        \begin{tablenotes}
		    \item[*] Implemented by ourselves.
		\end{tablenotes}
    \end{threeparttable}
    \end{adjustbox}
	\vspace{-0.35cm}
\end{table*}

\begin{table}[!t]
    \centering
    \caption{Results of our approach compared with different methods on the Colored MNIST dataset(mean $\pm$ std deviation).} 
    \label{table:cmnist}
	\vspace{-0.3cm}
	\begin{adjustbox}{max width=0.46\textwidth}
	\begin{threeparttable}
        \begin{tabular}{ll}
            \toprule
            \toprule
            Model & Acc test env\\
            \midrule
            ERM\tnote{*}  &  17.10 $\pm$ 0.6 \\
            IRM~\cite{arjovsky2019invariant} &  66.90 $\pm$ 2.5 \\
            Rex~\cite{krueger2020outofdistribution} &  68.70 $\pm$ 0.9 \\
            F-IRMGames~\cite{ahuja2020invariant}  & 59.91 $\pm$ 2.7 \\
            V-IRMGames~\cite{ahuja2020invariant} & 49.06 $\pm$ 3.4 \\
            ReBias~\cite{bahng2019rebias}\tnote{*} & 29.40 $\pm$ 0.3\\
            JiGen~\cite{carlucci2019domain}\tnote{*}  &11.91 $\pm$ 0.4\\
            \midrule
            \emph{DecAug} & \emph{\textbf{69.60 $\pm$ 2.0}}\\
            ERM, grayscale model(oracle) &  73.00 $\pm$ 0.4  \\
            Optimal invariant model (hypothetical) &75.00 \\
            \bottomrule
            \bottomrule
        \end{tabular}
        \begin{tablenotes}
		    \item[*] Implemented by ourselves.
		\end{tablenotes}
    \end{threeparttable}
    \end{adjustbox}
	\vspace{-0.35cm}
\end{table}

\noindent \textbf{Illustrative Results on the Colored MNIST Dataset.} 
As shown in Table~\ref{table:cmnist}, DecAug achieves the best generalization performance on Colored MNIST, followed by risk regularization methods, such as Rex and IRM. For typical domain generalization methods, such as JiGen, and the recently proposed method ReBias, are misled by the spurious correlation existing in the training datasets. Notice that DecAug's performance is very close to ERM trained on grayscale MNIST which provides an upper bound for MLP to generalize on this task. As mentioned in \cite{arjovsky2019invariant}, typical domain generalization methods can only deal with one dimension of OoD problem where image style differs. Our method further improves the performance in Colored MNIST by decomposition and semantic augmentation in the feature space, which disregards spurious features that are correlated but not causal for predicting category. 

\begin{table}[!t]
    \centering
    \caption{Results of our method compared with different models on the NICO dataset.}
    \label{table:nico}
	\vspace{-0.3cm}
	\begin{adjustbox}{max width=0.55\textwidth}
	\begin{threeparttable}
        \begin{tabular}{lcr}
        \toprule
        \toprule
        Model & Animal &  Vehicle \\
        \midrule
        ERM\tnote{*}  & 75.87 & 74.52 \\
        IRM~\cite{ahuja2020invariant}\tnote{*}  &59.17&62.00   \\
        Rex~\cite{krueger2020outofdistribution}\tnote{*} & 74.31 & 66.20 \\
        Cumix~\cite{mancini2020towards}\tnote{*} 
        &76.78 &74.74 \\
        DANN~\cite{ganin2016domain}\tnote{*}
        &75.59 &72.23 \\
        JiGen~\cite{carlucci2019domain}\tnote{*} &  84.95& 79.45 \\
        CNBB~\cite{he2020towards}\tnote{*} &78.16& 77.39	 \\
        \midrule
        \emph{DecAug} & \emph{\textbf{85.23}} & \emph{\textbf{80.12}} \\
        \bottomrule
        \bottomrule
        \end{tabular}
        \begin{tablenotes}
		    \item[*] Implemented by ourselves.
		\end{tablenotes}
     \end{threeparttable}
     \end{adjustbox}
	\vspace{-0.35cm}
\end{table}

\begin{figure*}
    \centering
    \includegraphics[width=0.85\linewidth]{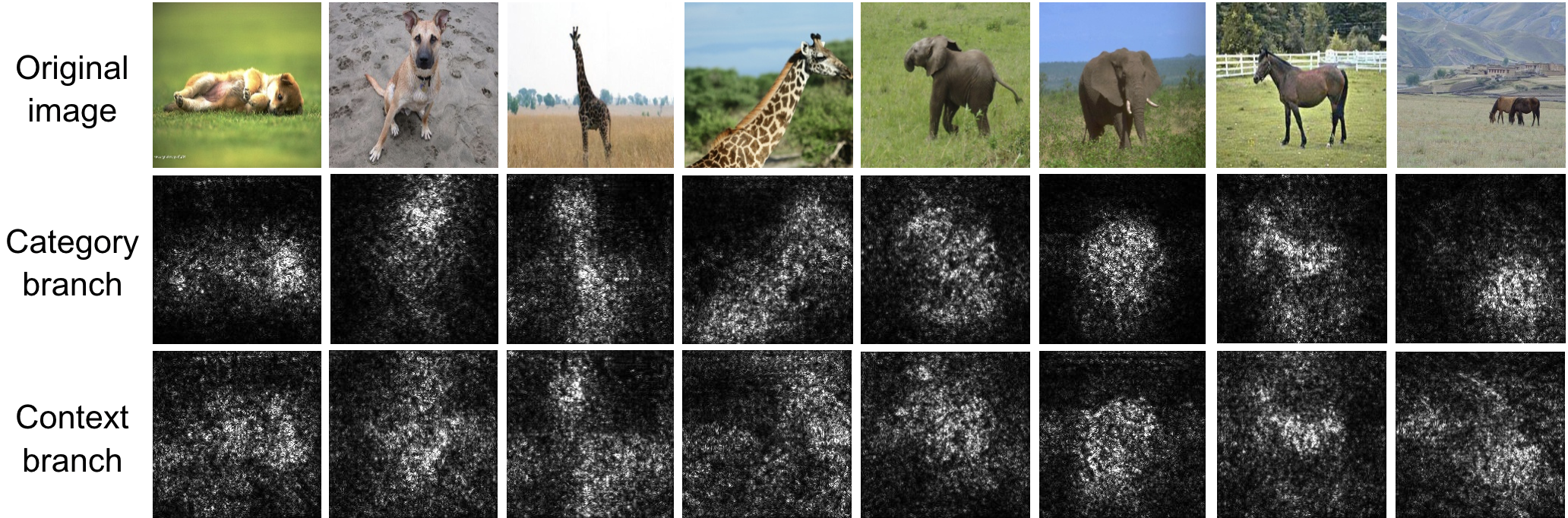}
    \caption{The gradient visualization of the decomposed category-related and context-related high-dimensional features. The first row is the original input images, the second row is its corresponding back propagation of the category branch and the last row is the back propagation of the context branch.}
    \label{fig:visualize}
    \vspace{-0.3cm}
\end{figure*}

\noindent \textbf{Illustrative Results on the PACS Dataset.}
In PACS, DecAug achieves the \textbf{\emph{state-of-the-art (SOTA)}} performance followed by MASF and Cumix when using ResNet-18 as the backbone network. The details of our results on PACS are shown in Table~\ref{table:pacs}. The worse performances for risk regularization methods, such as IRM and Rex, are because these methods add strong regularization terms in ERM to eliminate all features that are unstable across different environments. This can work well in the standard and ``clean" dataset—-MNIST, where shapes of digits are always stable. However, in realistic scenarios, the shapes of target objects can change, meaning that even features for predicting object category can be unstable in different training environments. 

\noindent \textbf{Illustrative Results on the NICO Dataset.}
The recently proposed NICO dataset, considers more realistic generalization scenarios, where objects themselves and the backgrounds \ie contexts in the dataset, can change vastly. For example, in the training dataset, we have dog pictures on the grass with dog faces posed to the camera, while in the test dataset, there are pictures where dogs are moving on the beachside. 
In our implementation, CuMix achieved 76.78\% (animal) and 74.74\% (vehicle) accuracy on NICO, indicating that mixing up (interpolating) data may not able to correct the spurious correlation between irrelevant features such as the background to the predicted category. In addition, DANN achieved 64.77\% (animal) and 58.16\% (vehicle) accuracy, which is similar to IRM. The poor performance of DANN and IRM on NICO may probably due to the diversity shift.
As Table~\ref{table:nico} shows, the proposed DecAug achieved the best generalization performances on two sets, followed by JiGen. This further demonstrates the superiority of the proposed algorithmic framework. \emph{Our method has achieved the SOTA performance simultaneously on various OoD generalization tasks, indicating a new promising direction for OoD learning algorithm research}.

\begin{table}[t]
    \centering
    \caption{Ablation study on PACS with ResNet-18.}
    \label{table:pacs_ablation}
	\vspace{-0.2cm}
	\begin{adjustbox}{max width=0.47\textwidth}
        \begin{tabular}{lcccc|c}
        \toprule
        \toprule
         PACS &  Art Painting &   Cartoon &  Sketch  &  Photo  &  Average\\
        \midrule
        DecAug without orth loss  &  78.42	 & 78.32 & 	72.13 & 94.19 &  80.77\\
        DecAug (orth 0.0005) & 77.49 & 77.43 & 74.32 & 94.07& 80.76\\
        DecAug (orth 0.001) & 78.12 & 77.34 & 76.97 & 94.91 & 81.83\\
        DecAug (orth 0.01) &  \emph{\textbf{79.00}} & \emph{\textbf{79.61}} & 	\emph{\textbf{75.64}} & 	\emph{\textbf{95.33}} &  \emph{\textbf{82.39}}\\
        \bottomrule
        \bottomrule
        \end{tabular}
    \end{adjustbox}
    \vspace{-0.3cm}
\end{table}

\subsection{Ablation Studies and Sensitivity Analysis}
For ablation studies and sensitivity analysis, we take the PACS dataset for example. 

\noindent\textbf{Effectiveness of orthogonal loss.}
We test the effects of the proposed orthogonal loss. The results are shown in Table~\ref{table:pacs_ablation}. It can be seen that without the orthogonal loss, our method achieves the average accuracy of 80.77\% that is higher than most of the methods in Table~\ref{table:pacs}. This is because the category and context losses also play the role of feature decomposition. The additional orthogonal loss enforces the gradients of the category and context losses to be orthogonal to each other, which helps to further decompose the features. As expected, with the increase of the orthogonal regularization coefficient $\lambda^{\text{orth}}$, the performance of DecAug can be improved. The experimental results confirm the effectiveness of the proposed orthogonal loss.

\noindent\textbf{DecAug variants.}
We changed current orth regularization to orth constrains between features, refer to Table~\ref{table:orthfeatures}, which reaches 79.90\% on PACS, lower than the original DecAug. We also tried confusion regularization, as discussed in recent literature Open Compound Domain Adaptation~\cite{liu2020open}. It seems natural to incorporate confusion regularization into our method for better decomposition. However, after many trials, no improvements were observed.
We tried DecAug with DANN adversarial loss "orth" on PACS. As shown in Table~\ref{table:orthfeatures}, the result is around 81\%, lower than the original. This shows both gradient orthogonalization and semantic augmentation are both indispensable parts in the algorithm. We tried "adversarial augmentation" to Jigsaw, the result is much lower than Jigsaw. This shows that the two branch architecture is needed and adversarial augmentation is better performed on the context predicting branch to improve OoD generalization via challenging neural networks to unseen context information.

\begin{table}[!t]
    \centering
    \caption{Results of DecAug variants on the PACS dataset.}
    \label{table:orthfeatures}
	\vspace{-0.2cm}
	\begin{adjustbox}{max width=0.55\textwidth}
	\begin{threeparttable}
        \begin{tabular}{lc}
        \toprule
        \toprule
        Model & Average  \\
        \midrule
        DecAug (DANN loss) & 81.00  \\
        DecAug (orth between features) & 79.90  \\
        DecAug (gradient-based orth)& \emph{\textbf{82.39}}  \\
        \bottomrule
        \bottomrule
        \end{tabular}
     \end{threeparttable}
     \end{adjustbox}
	\vspace{-0.3cm}
\end{table}

\noindent\textbf{Interpretability analysis.}
We also use deep neural network interpretability methods in \cite{Bengio2018_sanity} to explain the neural network's classification decisions as shown in Figure~\ref{fig:visualize}. It can be seen that the saliency maps of the category branch focus more on foreground objects, while the saliency maps of the context branch are also sensitive to background contexts that contain domain information. 
This shows that our method well decomposes the high-level representations into two features that contain category and context information respectively. Later, by performing semantic augmentation on context-related features, our model breaks the inherent relationship between contexts and category labels and generalizes to unseen combinations of foregrounds and backgrounds.

\section{Conclusions}
\label{sec:conc}
In this paper, we propose \name{}, a novel decomposed feature representation and semantic augmentation method for various OoD generalization tasks. High-level representations for the input data are decomposed into category-related and context-related features to deal with the diversity shift between training and test data. Gradient-based semantic augmentation is then performed on the context-related features to break the spurious correlation between context features and image categories. To the best of our knowledge, this is the first method that can simultaneously achieve the SOTA performance on various OoD generalization tasks from different research areas, indicating a new research direction for OoD generalization research. For future work, we will construct a large OoD dataset from the industry to further improve the algorithm and put it into real practice.

\section{Acknowledgements}
Nanyang Ye was supported in part by National Key R\&D Program of China  2017YFB1003000, in part by National Natural Science Foundation of China under Grant (No. 61672342, 61671478, 61532012, 61822206, 61832013,  61960206002, 62041205), in part by the Science and Technology Innovation Program of Shanghai (Grant 18XD1401800, 18510761200), in part by Shanghai Key Laboratory of Scalable Computing and Systems.

{\small
\bibliography{reference}
}

\clearpage

\maketitle

\appendix

\begin{spacing}{1.5}

\section{Supplementary Materials}

\subsection{Ablation Study for Semantic Augmentation}

In DecAug, semantic augmentation is proposed to further alleviate the influence of spurious correlation. We do not use random noise but adversarial augmentation on the context branch to improve the generalization ability of the model. In this section, we investigate the effects of semantic augmentation. We follow the same leave-one-domain-out validation protocol and the same training, validation and test split as in~\citet{Li2017}. The maximum number of training epochs in our experiment is 100. We use the SGD optimizer with a batch size of 64. 
Recall that $\epsilon$ is the magnitude of perturbation in augmentation.
We test different values of $\epsilon$ when all the other hyper-parameters remain the same. The results are summarized in Table~\ref{table:aug_ablation}. The first row lists the performance of the baseline model without augmentation. As can be seen in rows 2 and 4, semantic augmentation significantly improves classification accuracy. DecAug without semantic augmentation achieves $79.64\%$ accuracy, which is lower than DecAug. This is because semantic augmentation uses adversarial perturbation on the semantic space for improving the robustness of DecAug to unseen semantic domains. Moreover, the magnitude of the perturbation has a great influence on the results. Inappropriate magnitudes of perturbation would lead to very poor performance. 

\begin{table}[h]
    \centering
    \caption{Ablation study for semantic augmentation on PACS with ResNet-18.}
    \label{table:aug_ablation}
	\begin{adjustbox}{max width=0.47\textwidth}
        \begin{tabular}{lcccc|c}
        \toprule
        \toprule
         PACS &  Art Painting &   Cartoon &  Sketch  &  Photo  &  Average\\
        \midrule
        DecAug without perturbation  &  75.83	 & 77.37 & 	70.81 & 94.55 &  79.64\\
        DecAug ($\epsilon = 1$) & \textbf{79.00} & \textbf{79.61} & \textbf{75.64} & \textbf{95.33} & \textbf{82.39}\\
        \hline
        DecAug ($\epsilon = 0.01$) & 75.24 & 77.65 & 69.89 & 94.67& 79.36\\
        DecAug ($\epsilon = 10$) & 74.39 & 77.47 & 	75.24 & 	95.09 &  80.55\\
        DecAug ($\epsilon = 100$) & 63.13 & 70.39 & 64.19 & 94.55 & 73.07\\
        DecAug ($\epsilon = 1000$) & 62.89 & 71.97 & 71.44 & 41.13 & 61.86 \\
        \bottomrule
        \bottomrule
        \end{tabular}
    \end{adjustbox}
\end{table}

\subsection{Ablation Study for Features Concatenation}

In this section, we conduct an ablation study for features concatenation. Following the same leave-one-domain-out validation protocol as in the previous works on PACS, we evaluate the proposed DecAug without the concatenate operation. 
The balance weight for the category branch and the context branch is one. Note that we also involve the orthogonal constraints in this experiment.
As can be seen in Table~\ref{table:concat_ablation}, the application of features concatenation achieves the best performance.

\begin{table}[!h]
    \centering
    \caption{Ablation study for features concatenate on PACS with ResNet-18.}
    \label{table:concat_ablation}
	\begin{adjustbox}{max width=0.47\textwidth}
        \begin{tabular}{lcccc|c}
        \toprule
        \toprule
         PACS &  Art Painting &   Cartoon &  Sketch  &  Photo  &  Average\\
        \midrule
        DecAug with concat & \textbf{79.00} & \textbf{79.61} & \textbf{75.64} & \textbf{95.33} & \textbf{82.39} \\
        DecAug without concat & 75.83 & 76.24 & 75.57 & 94.79 & 80.61 \\
        \bottomrule
        \bottomrule
        \end{tabular}
    \end{adjustbox}
\end{table}

\subsection{Gradient Visualization of DecAug in NICO}

\begin{figure*}[bp]
    \centering
    \subfigure[Animal]{\includegraphics[width=0.9\linewidth]{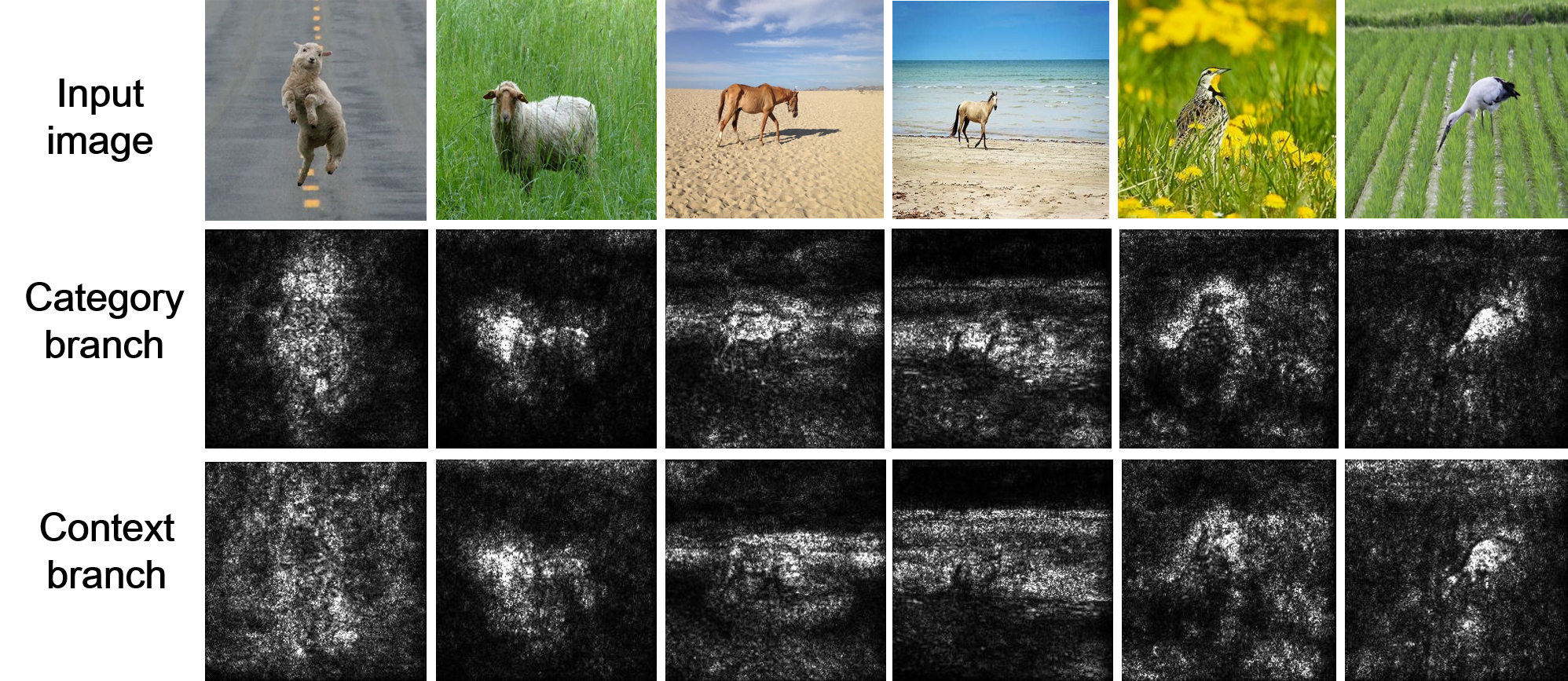}}
    \subfigure[Vehicle]{\includegraphics[width=0.9\linewidth]{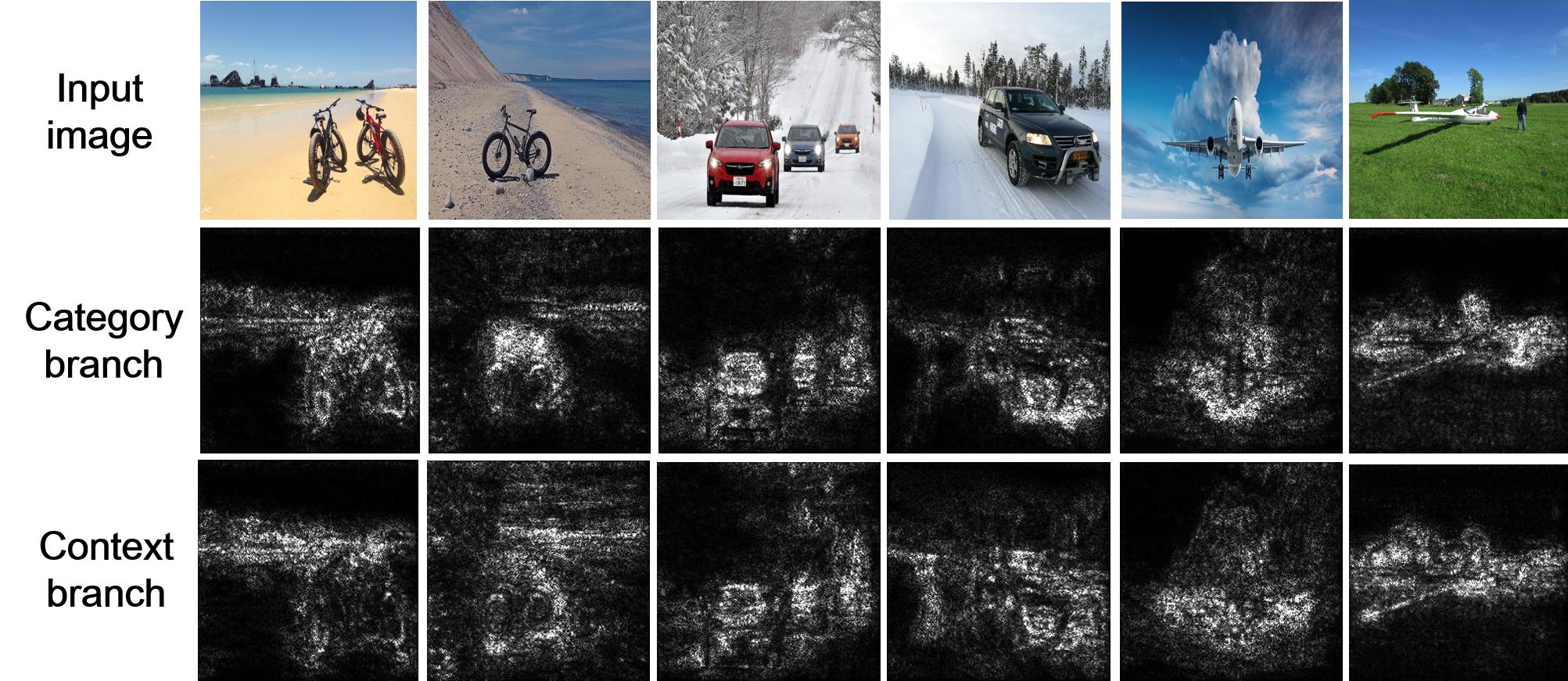}}
    \caption{The gradient visualization of the decomposed category-related and context-related high-dimensional features in NICO data. The first row is the original input images, the second row is its corresponding back propagation of the category branch and the last row is the back propagation of the context branch.}
    \label{fig:visualize_nico}
    \vspace{-0.3cm}
\end{figure*}

We use deep neural network interpretation methods to explain the neural network’s classification decisions in NICO, as shown in Figure~\ref{fig:visualize_nico}. 
It can be seen that the saliency maps of the category branch focus more on foreground objects, while the saliency maps of the context branch are also sensitive to background contexts that contain domain information.

\subsection{Experimental Details of NICO}

The NICO dataset contains two superclasses: animal and vehicle, which is a newly proposed OoD generalization benchmark in the real world. The original NICO dataset describes different ways to split training, validation and test set. For a fair comparison, we fix the split settings and constrain that the contexts have no intersection among the training, validation and test sets. The details of the split principle used in our experiments are shown in Table~\ref{table:split_animal} and Table~\ref{table:split_vehicle}.

\end{spacing}

\begin{table*}[!t]
    \centering
    \caption{NICO Dataset split in our experiment of each context for every class in Animal superclass.}
    \label{table:split_animal}
	\begin{adjustbox}{max width=0.80\textwidth}
        \begin{tabular}{llcccccccccc}
        \toprule
        \toprule
        Dataset split &Animal &Bear &Bird &Cat &Cow &Dog &Elephant &Horse &Monkey &Rat &Sheep \\
        \midrule
        domain 1 &on ground &$\surd$ &$\surd$ & & & & & & & & \\
        ~        &on snow   &$\surd$ & &$\surd$ &$\surd$ &$\surd$ &$\surd$ &$\surd$ &$\surd$ &$\surd$ &$\surd$ \\
        ~        &on grass  & &$\surd$ &$\surd$ &$\surd$ &$\surd$ &$\surd$ &$\surd$ &$\surd$ &$\surd$ &$\surd$ \\
        ~        &aside people & & & &$\surd$ & & &$\surd$ & & &$\surd$ \\
        \midrule
        domain 2 &eating grass &$\surd$ & & & & & & & & & \\
        ~        &in forest    &$\surd$ & & &$\surd$ & &$\surd$ &$\surd$ &$\surd$ &$\surd$ &$\surd$ \\
        ~        &in water     &$\surd$ &$\surd$ &$\surd$ & & $\surd$ & & &$\surd$ &$\surd$ &$\surd$ \\
        ~        &eating       & & $\surd$ &$\surd$ &$\surd$ &$\surd$ &$\surd$ & & $\surd$ &$\surd$ &$\surd$ \\
        ~        &on branch    & &$\surd$ & & & & & & & & \\
        ~        &in river     & & &$\surd$ &$\surd$ & & $\surd$ &$\surd$ & & & \\
        \midrule
        domain 3 &black &$\surd$ & & & & & & & & & \\
        ~        &brown &$\surd$ & & & & & & & & & \\
        ~        &lying &$\surd$ & & &$\surd$ &$\surd$ &$\surd$ &$\surd$ & & $\surd$ &$\surd$ \\
        ~        &flying& &$\surd$ & & & & & & & & \\
        ~        &in cage & &$\surd$ &$\surd$ & &$\surd$ & & &$\surd$ &$\surd$ & \\
        ~        &standing & & $\surd$ & & & &$\surd$ & & & & \\
        ~        &at home & & &$\surd$ &$\surd$ &$\surd$ & &$\surd$ & &$\surd$ & \\
        ~        &walking & & &$\surd$ & & & & & $\surd$ & & $\surd$\\
        ~        &on beach & & & & &$\surd$ & &$\surd$ &$\surd$ & & \\
        ~        &in zoo & & & & & &$\surd$ & & & & \\
        \midrule
        validation &on tree &$\surd$ & &$\surd$ & & & & & & & \\
        ~          &on shoulder &$\surd$ & & & & & & & & \\
        ~          &spotted & & & &$\surd$ & & & & & & \\
        ~          &running & & & & &$\surd$ & &$\surd$ & &$\surd$ & \\
        ~          &in circus & & & & & &$\surd$ & & & & \\
        ~          &climbing & & & & & & & &$\surd$ & & \\
        ~          &at sunset & & & & & & & & & &$\surd$ \\
        \midrule
        test       &white &$\surd$ & & & & & & & & & \\
        ~          &in hand & &$\surd$ & & & & & & & & \\
        ~          &in street & & &$\surd$ & &$\surd$ &$\surd$ &$\surd$ & & & \\
        ~          &standing & & & &$\surd$ & & & & & & \\
        ~          &sitting & & & & & & & &$\surd$ & & \\
        ~          &in hole & & & & & & & & &$\surd$ & \\
        ~          &on road & & & & & & & & & & $\surd$ \\
        \bottomrule
        \bottomrule
        \end{tabular}
    \end{adjustbox}
\end{table*}

\begin{table*}[!t]
    \centering
    \caption{NICO Dataset split in our experiment of each context for every class in Vehicle superclass.}
    \label{table:split_vehicle}
	\begin{adjustbox}{max width=0.80\textwidth}
        \begin{tabular}{llccccccccc}
        \toprule
        \toprule
        Dataset split &Vehicle &Airplane &Bicycle &Boat &Bus &Car &Helicopter &Motorcycle &Train & Truck \\
        \midrule
        domain 1 &around cloud &$\surd$ & & & & & & & & \\
        ~        &on beach     &$\surd$ &$\surd$ &$\surd$ & &$\surd$ &$\surd$ &$\surd$ &$\surd$ &$\surd$ \\
        ~        &in sunset    & &$\surd$ &$\surd$ & &$\surd$ &$\surd$ &$\surd$ &$\surd$ &$\surd$ \\
        ~        &double decker & & & &$\surd$ & & & & & \\
        ~        &on bridge     & & & &$\surd$ &$\surd$ & & &$\surd$ &$\surd$ \\
        \midrule
        domain 2 &aside mountain &$\surd$ & & & & &$\surd$ & &$\surd$ &$\surd$ \\
        ~        &at airport & & & & & & & & \\
        ~        &in city    &$\surd$ & &$\surd$ &$\surd$ &$\surd$ &$\surd$ &$\surd$ & &$\surd$ \\
        ~        &on snow & &$\surd$ & &$\surd$ &$\surd$ &$\surd$ &$\surd$ &$\surd$ &$\surd$ \\
        ~        &shared  & &$\surd$ & & & & & & & \\
        ~        &with people & &$\surd$ &$\surd$ &$\surd$ &$\surd$ &$\surd$ &$\surd$ & & \\
        ~        &in river & & &$\surd$ & & & & & & \\
        ~        &cross tunnel & & & & & & & &$\surd$ & \\
        \midrule
        domain 3 &at night &$\surd$ & & & & & & & & \\
        ~        &on grass &$\surd$ &$\surd$ & & & &$\surd$ &$\surd$ & &$\surd$ \\
        ~        &taking off &$\surd$ & & & & & & & \\
        ~        &in street & &$\surd$ & & & & &$\surd$ & & \\
        ~        &on road   & &$\surd$ & & &$\surd$ & &$\surd$ & &$\surd$ \\
        ~        &sailboat & & &$\surd$ & & & & & & \\
        ~        &wooden & & &$\surd$ & & & & & & \\
        ~        &yacht & & &$\surd$ & & & & & & \\
        ~        &aside traffic light & & & &$\surd$ & & & & & \\
        ~        &at station & & & &$\surd$ & & & &$\surd$ & \\
        ~        &on booth & & & & &$\surd$ & & & & \\
        ~        &at heliport & & & & & &$\surd$ & & & \\
        \midrule
        validation &in sunset &$\surd$ & & & & & & & & \\
        ~          &in garage & &$\surd$ & & & & &$\surd$ & & \\
        ~          &cross bridge & & &$\surd$ & & & & & & \\
        ~          &at yard & & & &$\surd$ & & & & & \\
        ~          &at park & & & & &$\surd$ & & & & \\
        ~          &on sea & & & & & &$\surd$ & & & \\
        ~          &subway & & & & & & & &$\surd$ & \\
        ~          &in race & & & & & & & & &$\surd$ \\
        \midrule
        test       &with pilot &$\surd$ & & & & & & & & \\
        ~          &velodrome  & &$\surd$ & & & & & & & \\
        ~          &at wharf & & &$\surd$ & & & & & & \\
        ~          &aside tree & & & &$\surd$ & & & & & \\
        ~          &on track & & & & &$\surd$ & &$\surd$ & & \\
        ~          &in forest & & & & & &$\surd$ & &$\surd$ &$\surd$ \\
        \bottomrule
        \bottomrule
        \end{tabular}
    \end{adjustbox}
\end{table*}

\end{document}